\pdfoutput=1

\documentclass[11pt]{article}

\usepackage{acl}

\usepackage{times}
\usepackage{latexsym}
\usepackage{enumitem}
\usepackage{comment}
\usepackage{multirow}
\usepackage{graphicx}
\usepackage{amsmath}
\usepackage{amssymb}
\usepackage{float}
\usepackage{natbib}

\usepackage[T1]{fontenc}

\usepackage[utf8]{inputenc}

\usepackage{microtype}

\title{Exploring Generative Models for Joint \\Attribute Value Extraction from Product Titles}

\author{Kalyani Roy \\
  IIT Kharagpur, India \\
  kroy@iitkgp.ac.in\\ \And
  Tapas Nayak\thanks{\hspace{2mm}This work was carried out while author was a postdoctoral researcher at IIT Kharagpur.} \\
  TCS Research, India \\
  tnk02.05@gmail.com \\ \And
  Pawan Goyal \\
  IIT Kharagpur, India \\
  pawang@cse.iitkgp.ac.in}

\begin{document}
\maketitle
\begin{abstract}

Attribute values of the products are an essential component in any e-commerce platform. Attribute Value Extraction (AVE) deals with extracting the attributes of a product and their values from its title or description. In this paper, we propose to tackle the AVE task using generative frameworks. We present two types of generative paradigms, namely, word sequence-based and positional sequence-based, by formulating the AVE task as a generation problem. We conduct experiments on two datasets where the generative approaches achieve the new state-of-the-art results. This shows that we can use the proposed framework for AVE tasks without additional tagging or task-specific model design.

\end{abstract}

%
%

\section{Introduction}
Product attribute values are crucial in any e-commerce platform. These provide the details of a product and help customers search for a product and make a purchase decision. Generally, the product title and description is carefully created to represent the main feature of a product, but the attribute-values may not always be explicit. Attribute Value Extraction (AVE) refers to the task of finding the values of attributes from the product title or description. For example, given the title, ``Seiko Women's SUJ708 Gold Tone Stainless Steel Watch'', its values can be \textit{SUJ708}, \textit{Seiko}, \textit{Gold}  and the associated attributes for these values can be \textit{Model number}, \textit{Brand}, and \textit{Band color}, respectively.



Most existing studies on AVE use neural sequence labeling architectures~\citep{hu-2011,OpenTag_2018,xu-etal-2019-scaling,TXtract_2020}. They use a set of entity-tags for each attribute (e.g., ``B-Brand” and ``I-Brand” for the attribute ``Brand''). 
To extract multiple attribute values, one line of work develops one model per attribute. 
However, multiple attribute-specific models might not be suitable when we have a large set of attributes. Instead of creating separate models, one can also use different tag sets corresponding to different attributes. ~\citet{xu-etal-2019-scaling} regards attributes as queries and adopts a global BIO tag for all the attributes, making it scalable for large attribute sets.
Recent works~\cite{xu-etal-2019-scaling, AVEQA_2020} formulate the attribute value extraction as a
question answering task, and their main goal is to extract the corresponding values for a given attribute.
~\citet{adatag_2021} proposes an adaptive decoding strategy to handle value extraction as a sequence tagging task. 
~\citet{JAVE-2020} introduces a multi-modal network that combines textual and visual information 
for joint attribute prediction and value extraction. ~\citet{roy-2021} formulates the task as a text infilling task and as an answer generation task and employ language models. 
In this paper, we attempt to jointly generate the values and their corresponding attribute names, whereas ~\citet{roy-2021, xu-etal-2019-scaling, AVEQA_2020} requires the attribute names as input to generate their values.


Motivated by recent success in formulating several NLP problems such as named entity recognition, question answering, sentence completion, etc.~\cite{t5_2020} as text generation problems, we propose to tackle the problem of attribute value extraction in a generative approach without any task-specific model designs. 
We present two paradigms, namely, word sequence-based and positional sequence-based, to transform the original task into a generative framework. The two formulations are detailed in Section~\ref{sec:gen_paradigm}. 

We summarize our contributions as follows: (1) To the best of our knowledge; we are the first to tackle the joint attribute value extraction using generative approaches. (2) We propose two generative paradigms to formulate this task as a generation problem. 
(3) We conduct experiments on two datasets, and the generative approaches surpass previous tagging-based methods. 

\section{Generative AVE}
\subsection{AVE with Generative Paradigm} \label{sec:gen_paradigm}

In this work, we tackle the task of product attribute value extraction in a generative manner. The input to the task is a ``product title'', and the outputs are the product attributes and
the corresponding values. To formulate the AVE task into a generative framework, we propose two paradigms: word sequence-based and positional sequence-based. \\

\begin{minipage}[h]{0.9\linewidth}
\texttt{Product Title} : New Band Women Skiing Jacket Outdoor Thicken Snowboarding Jacket Waterproof Windproof Outerwear Hooded Ski Coats WY006\\
\noindent
\texttt{Target (word sequence-based) :} Women ; Gender $\vert$ Snowboarding ; Sport Type $\vert$ Hooded ; Collar $\vert$ WY006 ; Model Number\\
\noindent
\texttt{Target (positional sequence-based): } \\2 2 Gender $\vert$ 7 7 Sport Type $\vert$ 12 12 Collar $\vert$ 15 15 Model Number \\ 
\end{minipage}

One example for these generative formulations is shown above. In both paradigms, each attribute-value pair is separated by a vertical bar. In the word sequence-based paradigm, we combine the value and the attribute with ``;''~\footnote{Any other separator symbols may be used to segregate the attribute values}  and in the positional sequence-based paradigm, we represent each attribute-value pair as a 3-point tuple - the start position and the end position of the value, and the attribute corresponding to that value~\footnote{While the value is explicitly present in the title, that is not always the case with the attribute.}. 

The positional sequence-based paradigm is close to the question-answer-based attribute-value extraction framework. The attribute is treated as the question, and the value is the answer to the question. However, our proposed formulation does not require the attribute to be given as the input to the model; it generates both the attribute and the span of the values. 

\subsection{Generative Models}
Given the input title sentence $T=\{t_1, t_2, t_3, ..., t_n\}$ with $n$ words, our goal is to generate a target sequence $Y$, which can either be word sequence-based or positional sequence-based. Let the text generation model be $f(\cdot)$. We can find the attribute-value pairs from $Y$. We separate the pairs with a vertical bar ``$|$". In the case of the word sequence paradigm, we get the attribute and the value by separating the sequence with ``$;$''. If two-word sequences are not isolated by ``$;$'', we can ignore them. In the case of the positional sequence-based paradigm, the first two tokens are the start index and the end index of the value, and the rest constitutes the attribute related to that value. If any of these three segments is missing in the generated text, we discard them. 

We adopt the \textbf{WDec}~\cite{aaai_Nayak_2020}, \textbf{PNDec}~\cite{aaai_Nayak_2020}, \textbf{BART}~\cite{bart_2020} and \textbf{T5}~\cite{t5_2020} as the generation model $f(\cdot)$. BART and T5 are transformer-based models that show superior performance in many seq2seq tasks. Both BART and T5 are used in both paradigms. We use WDec in the sequence-based paradigm and PNDec in the positional sequence-based paradigm. WDec employs seq2seq model to generate the word sequence. 
PNDec jointly extracts the value and attribute in the sentence with a pointer network-based decoder. We replace their encoder with BERT~\cite{bert}. By formulating the AVE task as a text generation problem, we can tackle the problem without any additional tagging or task-specific model design.

\section{Experiments}
\subsection{Dataset and Evaluation Metrics}

\textbf{Dataset:}\hspace{5mm}For our experiments, we derive two datasets from 1)~\citet{xu-etal-2019-scaling}~\footnote{\url{https://github.com/lanmanok/ACL19_Scaling_Up_Open_Tagging/blob/master/publish_data.txt}} and 2) Multimodal Attribute Extraction (MAE) by~\citet{MAE-AKBC_2017}~\footnote{\url{https://rloganiv.github.io/mae/}}. The dataset by ~\citet{xu-etal-2019-scaling} consists of tuples of title, attribute, and value from Sports \& Entertainment category of AliExpress~\footnote{\url{https://www.aliexpress.com/}}~\footnote{We could not use the dataset MEPAVE introduced by~\citet{JAVE-2020} as it is in Chinese language.}. We remove the tuple that has NULL in their value, i.e., the attribute's value is missing in that title. We keep only those attributes that appear at least $60$ times~\footnote{We empirically decide this threshold.} in the dataset. More than $67\%$ attributes appeared only twice in the dataset. These rarely occurring attributes did not help our generative models. So, we discard the less frequently occurring attributes. We combine the tuples having the same title. 
The input text is the title, and the attributes and the corresponding values are the expected output. We will denote this dataset as AV-DATA-V1. 

The MAE dataset is collected from many e-commerce sites and includes products from diverse categories such as electronic products, jewelry, clothing, vehicles, and real estate. It also consists of product title, description, and attribute-value pairs. This dataset is very noisy and needs extensive pre-processing and cleaning. Although this dataset has a rich set of attributes, we find that many of its values are ``yes'', ``no'', ``na'', which can not be obtained from the product title, and we discard those. We also remove the attributes whose values are not present in the title. 
Few of the attributes occur very frequently, like, colour, colour-family, etc., and some of the attributes occur very rarely. The attributes with high frequency overshadow the training. So, we discard those attributes. As these rare attributes are not helping the model in learning, we also discard those. We empirically consider only those attributes that occur at least $700$ times in the MAE dataset. We will refer to this dataset as AV-MAE. 

The datasets are randomly split into 80:10:10 ratio as train, validation and test set. Table~\ref{tab:dataset} shows the details of the datasets.  
Overall, there are $62$ and $95$ attributes in AV-DATA-V1 and AV-MAE respectively. We will share our datasets upon acceptance of this paper.

\noindent
\textbf{Evaluation Metrics:}\hspace{5mm}Following previous works~\cite{OpenTag_2018, xu-etal-2019-scaling, adatag_2021}, we use precision, recall, and F1 score as our evaluation metrics for this joint attribute value extraction task. We remove the duplicates from the extracted pairs. We consider a predicted pair correct if the corresponding attribute name and the value both exactly match the ground truth pairs. To get insight into the models' ability to identify the attributes and values separately, we also compute the precision, recall, and F1 score for the attributes and the values.

\begin{table}[t]
\centering
\small
\begin{tabular}{ccrrr}
\hline
Dataset   &        & \multicolumn{1}{c}{Train} 
& \multicolumn{1}{c}{Valid} & \multicolumn{1}{c}{Test} \\ \hline
\multirow{3}{*}{AV-DATA-V1} 
	& \#Sent & 28,348  & 3,544   & 3,544  \\
	& Single & 12,621  & 1,598   & 1,603  \\
    & Multi  & 15,727  & 1,946   & 1,941  \\ \hline
\multirow{3}{*}{AV-MAE} 
    & \#Sent & 78,402  & 9,694   & 9,690  \\
    & Single & 71,769  & 8,876   & 9,050  \\
    & Multi  & 6,633   & 818     & 640    \\ \hline
\end{tabular} 
    \caption{Statistics of AV-DATA-V1 and AV-MAE. \#Sent. represents the total number of sentences. Single and Multi respectively represent \# sentences with single and multiple attribute-value pairs. } 
    \label{tab:dataset}
\end{table}

\subsection{Experimental Setup}

We adopt \textbf{BiLSTM-CRF}~\cite{Huang2015BidirectionalLM}, \textbf{SUOT}~\cite{xu-etal-2019-scaling},  \textbf{JAVE}~\cite{JAVE-2020}, \textbf{BERT-NER}~\cite{bert} for tagging-based approaches. All these models use BERT~\cite{bert} to encode the sentences.

\noindent
(1) \textbf{BiLSTM-CRF} uses the word embedding from pretrained BERT model. It uses a BiLSTM-based encoder and a CRF-based decoder. (2) \textbf{SUOT} requires both the title and the attribute to predict the value. 
It applies cross attention between the title and the attribute, followed by a CRF layer. We feed the model with the title and all the possible attributes in the corpus. The model should not predict any values for the attributes not in the ground truth. 
(3) \textbf{JAVE} predicts the attributes using a classifier and then considers the value extraction as a sequence labeling task. (4) \textbf{BERT-NER} does not apply CRF to decode the tags. It feeds the representation of the first sub-token to the token-level classifier to predict the label.



\begin{table*}[th]
\centering
\resizebox{0.95\linewidth}{!}{
\begin{tabular}{@{\extracolsep{5pt}} l l l ccc ccc ccc @{}}
\hline
\multirow{2}{*}{Dataset}    & \multirow{2}{*}{Paradigm}     & \multirow{2}{*}{Model} & \multicolumn{3}{c}{Attribute+value} & \multicolumn{3}{c}{Attribute} & \multicolumn{3}{c}{Value} \\
\cline{4-6} \cline{7-9} \cline{10-12}
 &   & & P & R  & F1  & P    & R    & F1  & P   & R  & F1 \\ \hline 

\multirow{10}{*}{\parbox{.075\textwidth}{AV-DATA-V1}}  
 & \multirow{4}{*}{\parbox{.075\textwidth}{Tagging} } 
 & BiLSTM-CRF & 
 \textbf{65.06} & 35.75 & 46.14  
 & \textbf{73.56} & 38.66 & 50.68 
 &  \textbf{68.87} & 37.84 & 48.84 \\
 
 & & SUOT &  -  & -   & -  
 & - & - & - 
 &  25.34 & \textbf{61.59}  & 35.91 \\
 
 &  & JAVE & 42.69 & 29.37 & 34.80 
 & 70.36 & 53.17 & 60.57 
 & 42.21 & 30.63 & 35.50   \\
 
 & & BERT-NER 
 & 64.10 & \textbf{56.36} & \textbf{59.98} 
 & 70.63 & \textbf{62.10} & \textbf{66.09} 
 &  66.24 & 58.24 & \textbf{61.98} \\ \cline{2-12}
 
 & \multirow{3}{*}{\parbox{.075\textwidth}{ Word \\ Sequence}} 
 & WDec & 43.74 & 32.00 & 36.96 
 & 49.48 & 36.20 & 41.81 
 & \textbf{72.45} & 53.01 & 61.22 \\
 
 &  & BART & 60.41 & 53.89 & 56.96 
 & 65.58 & 58.51 & 61.84 
 & 63.17 & 56.36 & 59.57 \\
 
 &  & T5  
 &  \underline{\textbf{68.67}} & \textbf{60.50} & \underline{\textbf{64.32}} 
 & \textbf{72.75} & \underline{\textbf{64.08}} & \underline{\textbf{68.14}} 
 & 70.72 & \textbf{62.30} & \textbf{66.29}   \\ \cline{2-12}
 
 & \multirow{3}{*}{\parbox{.075\textwidth}{Positional \\sequence}}    
 & PNDec & \textbf{65.89} & 50.75 & 57.34 
 & \underline{\textbf{77.99}} & 60.08 &\textbf{ 67.87}
 & 70.53 & 54.33 & 61.38 \\
 
 &  & BART   & 64.67 & \underline{\textbf{61.85}} & \textbf{63.23} 
 & 66.24 & \textbf{63.35} & 64.76 
 & \underline{\textbf{75.43}} & \underline{\textbf{72.15}} & \underline{\textbf{73.75}} \\
 
 &  & T5  &  63.21 & 54.66 & 58.62 
 & 65.43 & 56.58 & 60.68 
 & 73.30 & 63.39 & 67.99   \\\hline 
\multirow{10}{*}{\parbox{.075\textwidth}{AV-MAE}}      & \multirow{4}{*}{\parbox{.08\textwidth}{Tagging}}      
 & BiLSTM-CRF 
 &  63.46 & 47.12 & 54.09  
 & 67.15 & 49.37 & 
 56.90 & 71.24 & 52.90 & 60.71 \\
 
 & & SUOT   &  -  & -   & -  
 & - & - & - &  6.51 &  75.26 &  11.98 \\
 
 &   & JAVE   &   48.53 & 31.18 & 37.97 
 & 64.75 & 48.22 & 55.27 
 & 48.53 & 31.18 & 37.97  \\
 

& & BERT-NER
& \textbf{65.45} & \textbf{66.81} & \textbf{66.12 }
& \textbf{67.20} & \textbf{68.59} & \textbf{67.89}
& \textbf{74.40} & \textbf{75.94} & \textbf{75.16} \\ \cline{2-12}

 & \multirow{3}{*}{\parbox{.075\textwidth}{ Word \\ sequence}} 
 & WDec  &  38.55 & 33.93 & 36.09 
 & 40.87 & 35.88 & 38.16 & 73.94 & 65.09 & 69.24   \\
 
 &  & BART &  64.28 & 65.35 & 64.81 
 & \textbf{73.01} & \textbf{74.23} & \textbf{73.62} 
 & 66.31 & 67.42 & 66.86  \\
 
 &  & T5  &   \textbf{66.45} & \underline{\textbf{68.44}} & \textbf{67.43} 
 & 67.82 & 69.86 & 68.82 
 & \underline{\textbf{77.22}} & \underline{\textbf{79.54}} & \underline{\textbf{78.36}}  \\ \cline{2-12}
 
 & \multirow{3}{*}{\parbox{.08\textwidth}{Positional \\sequence}}     
 & PNDec &   63.59 & 63.12 & 63.35 
 & 67.75 & 67.25 & 67.50 
 & \textbf{74.58} & \textbf{74.03} & \textbf{74.30}\\
 
 &  & BART     &  65.15 & 66.66 & 65.89 
 & 74.27 & 76.00 & 75.13 
 & 68.34 & 69.94 & 69.13   \\
 
 &  & T5  &   \underline{\textbf{66.81}} & \textbf{68.43} & \underline{\textbf{67.61}} 
 & \underline{\textbf{76.79}} & \underline{\textbf{78.66}} & \underline{\textbf{77.71}} 
 & 69.32 & 71.01 & 70.16 \\ \hline     
\end{tabular}}
\caption{Comparative results on the AV-DATA-V1 and AV-MAE datasets. Bold values represent the best scores in each paradigm, and the underlined values are the best result in each dataset.}
\label{tab:result}
\vspace{-4mm}
\end{table*}



\begin{table}[ht]
\small
\centering
\begin{tabular}{@{\extracolsep{5pt}} l cc cc @{}}
\hline
 \multirow{2}{*}{Model}  & \multicolumn{2}{c}{AV-DATA-V1} & \multicolumn{2}{c}{AV-MAE} \\ \cline{2-3} \cline{4-5}
    & Single   & Multi  & Single   & Multi   \\ \hline
BERT-NER & 45.07  & 65.33  & 63.70  & \textbf{85.40}   \\
BART (W.S.)    & 41.11  & 62.49  & 63.80  & 76.17   \\
T5 (W.S.)  & 47.66  & \textbf{69.61}  & 65.09  & 85.32   \\
BART (P.S.)  & \textbf{51.57}  & 66.50  & 64.55  & 75.64   \\
T5 (P.S.)  & 43.79  & 63.57  & \textbf{65.49}  & 83.93  \\ \hline 
\end{tabular}
    \caption{Comparison of F1 score of sentences having single and multiple values for AV-DATA-V1 and AV-MAE datasets. W.S. and P.S. denote word sequence-based and positional sequence-based paradigms, respectively. }
    \label{tab:analysis}
    \vspace{-4mm}
\end{table}

\section{Result and Discussion} 
Table~\ref{tab:result} presents the overall results using both the datasets in joint attribute value extraction, attribute identification, and value extraction. It may be possible that the model generates the attribute correctly, but the identified value is incorrect or vice-versa. The second column shows the models’ ability to generate the attributes without considering the values. The third column shows the models’ ability to identify the probable values without considering the associated attributes. We observe that the generative methods, based on either word sequence or positional sequence, reach state-of-the-art results in both datasets. We find that the result is almost the same even if we shuffle the order of {value: attribute} pairs while training for the word sequence-based paradigm.

For joint attribute value extraction, T5 outperforms the other models in the AV-MAE dataset, but the performance of both BART and T5 is close. In AV-DATA-V1, BART achieves the best F1 score. 
PNDec, BART, and T5 consistently outperform tagging based approaches. The PNDec shows comparable performance with BART and T5 on both the datsets. JAVE shows competitive result for finding only the attributes. SUOT shows low precision, but high recall. On closer examination of the extracted value of SUOT, we find that the model is generally predicting the same values for different attributes. 
We observe that the scores for joint attribute value extraction and value prediction are the same for JAVE in AV-MAE. This happened because whenever the model had extracted the correct value, its attribute was also correct. 
Most of the methods are better equipped to predict the attributes. The models are better at predicting the attributes or generating the values separately, but when we consider both the attributes and the values, the performance decreases. In Table~\ref{tab:analysis} we compare F1 scores of sentences having single and multiple values as described in Table~\ref{tab:dataset}. The generative methods perform better than the tagging based approach in identifying sentences having single value. When the sentences have multiple values, the performance of BERT-NER and T5 is competitive in AV-MAE dataset. 
This shows the effectiveness of the generative model.

None of the datasets have annotated all possible attribute values and we find that the models are generating attribute values apart from the ones given in the target list which might be correctly generated, but as it is not in our gold-standard data, it is eventually marked as incorrect. 
We also find that some of the attributes, e.g., Model Number, have incorrect values in the AV-DATA-V1. This affects the model's ability in extracting attribute and values. In both the datasets, the span of any attribute is contiguous, but it is possible that the value spans multiple different positions. Although the word sequence-based paradigm may be able to handle this situation, the positional sequence-based paradigm needs to be extended in such cases.


\section{Conclusion and Future Work}
In this paper, we tackle the AVE task in a generative framework by formulating it with a word sequence-based paradigm and a positional sequence-based paradigm. Extensive experiments on two datasets show that our proposed formulations achieve state-of-the-art results. Current research works treat AVE as a sequence labeling problem, and our work is an attempt at transforming the task into text generation problems. Designing more effective generation paradigms can be an interesting research problem for future work.


\bibliographystyle{acl_natbib}

\setcounter{table}{0}
\renewcommand{\thetable}{A.\arabic{table}}

\begin{table*}[th]
    \centering
    \begin{tabular}{p{0.22\linewidth}|p{0.75\linewidth}}
    \hline
{Input Title}   &  adidas superstar gold label, men's skateboarding  shoes, white, wrap abrasion lightweight breathable b34308  \\ \hline
Target : Word seq. & adidas ; brand name | b34308 ; model number \\
WDec  &   adidas ; model | skateboarding shoes ; model \\
BART  &   adidas ; brand name | breathable ; feature \\
T5    &   adidas ; brand name | breathable ; feature | b34308 ; model number \\ \hline
Target: Pos. seq. & 0 0 brand name | 12 12 model number \\
Pos. seq. for PNDec : & 0 1 brand name | 25 28 model number \\ 
PNDec  &   25 28 model number | 0 1 brand name \\
BART  &  0 0 brand name | 11 11 feature | 12 12 model number \\
T5    & 0 0 brand name | 11 11 feature | 12 12 model number \\ \hline
    \end{tabular}
    \caption{Example of generated sequences in both word sequence-based paradigm and positional sequence-based paradigm.}
    \label{tab:case}
\end{table*}

\newpage

\appendix

\section{Appendix}
\subsection{Implementation Details}

For all the tagging-based method we use We adopt the BART-base and T5-base model from \textit{huggingface} Transformer library~\footnote{\url{https://github.com/huggingface/transformers}} for our experiments. 
We train the models with a batch size of 16 with a learning rate $3e-4$. The model that performs the best on the validation dataset is used for testing. All our model variants are trained on Tesla P100-PCIE 16GB GPU.

\subsection{Case Study}

Table~\ref{tab:case} shows an example of generated attribute values for the given product title ``adidas superstar gold label, men's skateboarding  shoes, white, wrap abrasion lightweight breathable b34308''. The middle group shows the word sequence-based paradigm and the last group shows the positional sequence-based paradigm. In the word sequence-based paradigm, WDec incorrectly predicts the model as ``skateboarding shoes''; BART can not find the model number, but it predicts another attribute ``feature'' with ``breathable'' as the value; T5 correctly identifies two attribute values and predicts another new attribute. This predicted value is correct, but as it is not annotated, it is marked as incorrect at the time of evaluation. PNDec identifies both the attribute and the values in the positional sequence-based paradigm. Since we use the BERT tokenizer for PNDec, the positions are different for PNDec, and BART, T5. In this paradigm, both BART and T5 identify a new attribute ``feature'' with the value ``breathable''.

\end{document}